\documentclass[conference]{IEEEtran}
\IEEEoverridecommandlockouts

\usepackage{cite}
\usepackage{amsmath,amssymb,amsfonts}
\usepackage{algorithmic}
\usepackage{graphicx}
\usepackage{textcomp}
\usepackage{xcolor}
\usepackage{booktabs}
\usepackage{multirow}
\usepackage{placeins}
\usepackage{enumitem}
\usepackage{hyperref}
\usepackage{subcaption}
\def\BibTeX{{\rm B\kern-.05em{\sc i\kern-.025em b}\kern-.08em
    T\kern-.1667em\lower.7ex\hbox{E}\kern-.125emX}}
\begin{document}

\title{Uncertainty-Aware Action Segmentation for Human-to-Robot Assembly Teleoperation}

\author{
\textbf{Fnu Heman}$^{*1,2}$\thanks{$^{*}$These authors contributed equally to this work.
$^{1}$Mechanical Engineering, University of California, Riverside, CA 92521, USA.
$^{2}$Electrical and Computer Engineering, University of Miami, FL 33146, USA.}, \textbf{Yixuan Wang}$^{*1}$, \textbf{Kolin Xu}$^{*1}$, \textbf{Conner Wallace}$^{1}$, \textbf{John Dang}$^{1}$,\\
\textbf{Akhil Joshi}$^{1}$, \textbf{Vignesh Thondikulam}$^{1}$, \textbf{Jun Sheng}$^{1}$, \textbf{Pinhas Ben-Tzvi}$^{2}$, \textbf{Mingyu Cai}$^{1 \dagger}$
}

\maketitle

\begin{abstract}
In assisted teleoperation for human-robot collaboration, understanding operator behavior is essential for reliable robotic assistance during long-horizon manipulation and assembly tasks. Recognizing and segmenting actions provides a foundation for intention estimation and execution monitoring. However, collecting robot teleoperation demonstrations is costly and constrained by hardware availability, whereas human demonstrations are easier to collect and capture rich temporal structure. We propose an uncertainty-aware action segmentation framework for long-horizon tasks that transfers temporal knowledge from human demonstrations to robot demonstrations. The framework integrates: (1) cross-domain temporal learning through human-domain pretraining and robot-domain fine-tuning of a temporal convolutional network (MS-TCN++~\cite{li2020mstcnpp}) for frame-level action segmentation; (2) a reliability-aware correction method that identifies uncertain segments and uses sequence context to selectively revise their predictions; and (3) a conformal prediction module that provides frame-level prediction sets for uncertainty quantification and evaluates their empirical coverage. Experiments on toy-car assembly demonstrations spanning 22 action classes show that human-to-robot fine-tuning improves test Edit score (a measure of segment-sequence similarity to ground truth) from 55.23 to 72.51 and segmental F1@50 (F1 score at 50\% temporal overlap) from 10.97 to 16.86 using 9 robot demonstrations. On the held-out test set, where 68\% of predicted Human$\rightarrow$Robot segments are incorrect, the reliability analyzer detects them with 0.755 AUROC. Unrestricted correction reduces Edit from 72.51 to 67.86, whereas selective acceptance preserves it (73.45), so the gate acts mainly as a safeguard against harmful relabeling rather than as a source of large gains. Class-conditional conformal prediction attains 95.38\% empirical coverage at a 93\% target, but prediction sets still contain 17.41 of 22 classes on average. These results demonstrate the benefit of human temporal pretraining while highlighting the challenges of reliable sequence correction and uncertainty quantification under limited robot supervision. \noindent\textbf{Code and data:} \href{https://snazzy-kelpie-a9faf0.netlify.app/}{link}.
\end{abstract}

\begin{figure*}[t]
  \centering
  \includegraphics[width=\textwidth]{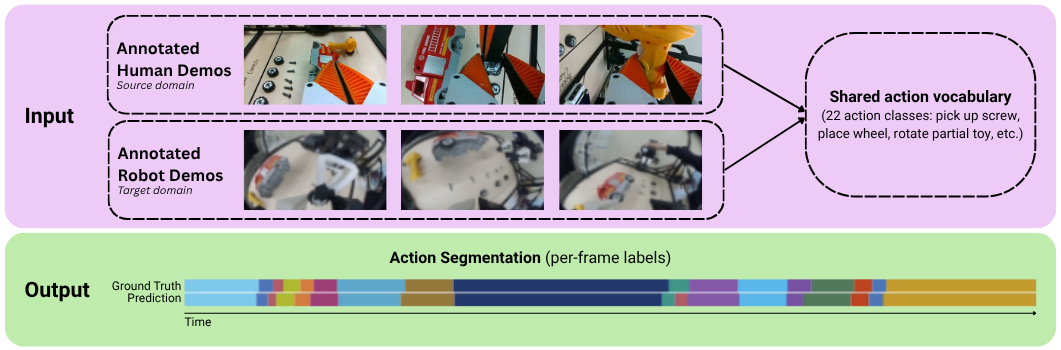}
    \caption{
    Human task-demonstration frames (top) and robot demonstration frames
    (bottom). The output sequence shows ground-truth (GT) and predicted
    (Pred) action segments. Colors denote action classes.
    }
  \label{fig:dataset_segmentation}
\end{figure*}
\section{Introduction}
During teleoperated assembly, actions such as picking up, aligning, and fastening a screw form long, continuous task sequences without explicit action boundaries. Segmenting these demonstrations into their constituent actions provides structured information for understanding task progression, learning from demonstrations, and analyzing execution errors. Temporal action segmentation (TAS) addresses this problem by assigning an action label to each frame of an untrimmed video. Methods such as MS-TCN++~\cite{li2020mstcnpp} and ASFormer~\cite{yi2021asformer} model temporal dependencies to produce coherent action sequences. However, adapting these models to robotic assembly requires annotated demonstrations that are costly to collect because of hardware availability, teleoperation effort, and annotation requirements.

Human demonstrations offer an accessible source of task information: a person directly manipulating a handheld gripper can demonstrate assembly procedures without operating a robot, while retaining action-ordering structure relevant to robot execution. Prior work has investigated intention estimation and task understanding in teleoperated assembly and human-robot collaboration~\cite{cai2024hierarchical,xu2025probabilistic,baskaran2025explainable,huang2023hierarchical}. Complementing these directions, we focus on fine-grained temporal action segmentation and investigate whether human-domain pretraining improves robot-domain performance when robot training data are limited, despite differences in appearance, embodiment, and execution timing.

Transfer alone does not resolve all segmentation errors. Incorrect action labels, fragmented predictions, and poorly localized transitions can persist after fine-tuning. Sequence-level context provides additional information for revising uncertain segments, but unconstrained correction can also overwrite useful predictions and degrade temporal segmentation. We therefore formulate correction as a detect--propose--select process that separates error detection, replacement generation, and selective acceptance.

We propose an uncertainty-aware human-to-robot action segmentation framework built on MS-TCN++~\cite{li2020mstcnpp}. Using fixed visual features, we pretrain the temporal model on human-operated gripper demonstrations and fine-tune it on robot teleoperation data, evaluating temporal model transfer without adapting the visual feature extractor. A reliability analyzer first flags potentially incorrect segments, a sequence-aware Transformer~\cite{vaswani2017attention} then uses predicted class probabilities and temporal context to propose replacement labels, and a gate fixed before test evaluation selectively accepts these proposals without ground-truth access.

We additionally integrate conformal prediction (CP)~\cite{DBLP:journals/corr/abs-2107-07511,stutz2021learning} to represent frame-level uncertainty through sets of plausible action labels. Standard split-conformal prediction provides a marginal coverage guarantee under exchangeability of calibration and test examples; because temporal dependence within videos complicates this assumption, we report empirical coverage together with prediction-set size, which together distinguish coverage from selectivity.

With only nine physical robot demonstrations for robot-domain training, the human-only model performs poorly zero-shot (13.40\% frame accuracy), while human-to-robot fine-tuning improves the mean test Edit score from 55.23 to 72.51 over robot-only training, evaluated across three random seeds on 17 held-out robot demonstrations. These findings demonstrate the value of human-domain pretraining for robot action segmentation under limited robot supervision.

Our main contributions are:
\begin{itemize}
\item We demonstrate that human-domain temporal pretraining improves robot action segmentation with only nine robot demonstrations, comparing human-only, robot-only, and human-to-robot MS-TCN++ models across three random seeds under fixed visual features.
\item We develop a reliability-aware detect--propose--select correction pipeline and show that unrestricted sequence-level relabeling degrades temporal segmentation, while selective acceptance prevents this degradation.
\item We quantify frame-level predictive uncertainty using standard and class-conditional split-conformal prediction, evaluating empirical coverage and mean prediction-set size before and after selective correction.
\end{itemize}

\section{Related Work}

\noindent\textbf{Learning from Human Demonstrations.}
Human demonstrations provide visual and temporal information for robot
learning. R3M~\cite{nair2022r3m} and MVP~\cite{radosavovic2023mvp} study
visual pretraining for downstream robot tasks, while Open
X-Embodiment~\cite{openx2023} studies learning across heterogeneous robot
datasets. UMI~\cite{chi2024umi} and ALOHA~\cite{zhao2023aloha} provide
complementary ways to collect human-operated gripper and robot
teleoperation demonstrations. We adopt this pairing with a shared action
vocabulary and hold visual features fixed to study transfer of the
temporal segmentation model.

\noindent\textbf{Temporal Action Segmentation and Transfer.}
Temporal convolutional networks~\cite{Lea_2017_CVPR},
MS-TCN~\cite{farha2019mstcn}, and MS-TCN++~\cite{li2020mstcnpp} model
temporal dependencies and refine frame-level predictions, while
ASFormer~\cite{yi2021asformer} and DiffAct~\cite{liu2023diffact} use
attention and diffusion, respectively. SSTDA~\cite{chen2020sstda}
addresses domain adaptation in TAS through unlabeled videos and
self-supervised temporal alignment. In contrast, we use supervised
fine-tuning on a small labeled robot dataset to study transfer across
embodiments.

Reducing annotation cost is a related but distinct direction:
timestamp-supervised TAS~\cite{li2021timestamp} uses sparse labels, and
multi-modal few-shot TAS~\cite{lu2025multimodal} generalizes to novel
procedural tasks from few examples. We instead keep a fixed vocabulary
and dense annotations. Common benchmarks include
50Salads~\cite{stein2013combining}, Breakfast~\cite{kuehne2014language},
GTEA~\cite{fathi2011learning}, and Assembly101~\cite{sener2022assembly101};
Assembly101 offers large-scale multi-view human assembly recordings,
whereas we evaluate human-gripper-to-robot-teleoperation transfer.

\noindent\textbf{Segment Refinement and Selective Correction.}
ASRF~\cite{ishikawa2021alleviating} combines action classification with
boundary regression to reduce over-segmentation. We instead refine
segments from a frozen backbone through a detect--propose--select
pipeline. Selective prediction offers a related view:
SelectiveNet~\cite{geifman2019selectivenet} jointly learns prediction and
rejection to manage the risk--coverage trade-off. Our gate instead
decides whether to apply a proposed correction, and rejection retains the
backbone prediction rather than abstaining.

\noindent\textbf{Uncertainty Quantification.}
Monte Carlo dropout~\cite{gal2016dropout} estimates predictive
uncertainty through stochastic forward passes. Conformal prediction
(CP)~\cite{vovk2005algorithmic,DBLP:journals/corr/abs-2107-07511}
constructs prediction sets with finite-sample marginal coverage under
exchangeability, and prior work studies set
efficiency~\cite{romano2020classification,angelopoulos2020uncertainty,stutz2021learning}
and distribution shift~\cite{tibshirani2019conformal}. Because temporal
dependence within videos complicates exchangeability, we report empirical
frame-level coverage together with prediction-set size.
\section{Method}

\begin{figure*}[t]
    \centering
    \includegraphics[width=\textwidth]{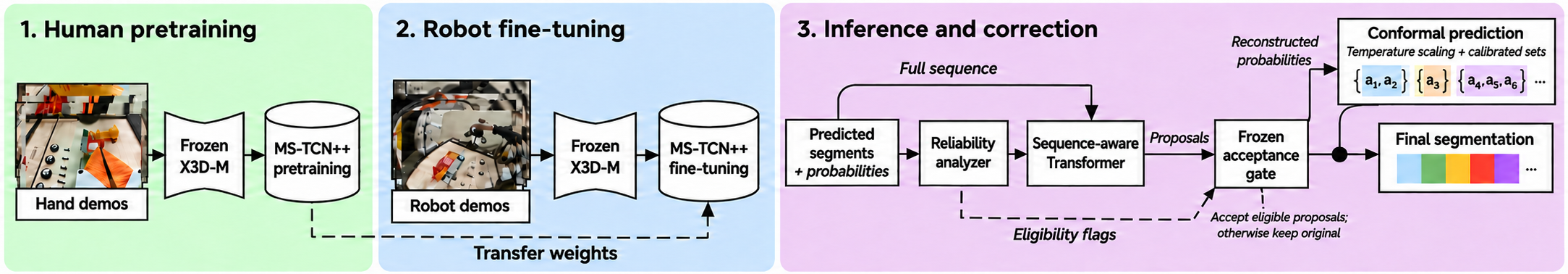}
    \caption{
    \textbf{Overview of the proposed framework.}
    (1) Human pretraining of MS-TCN++, (2) robot-domain fine-tuning through
    transferred weights, and (3) reliability-aware selective sequence correction
    with conformal prediction for frame-level uncertainty quantification.
    }
    \label{fig:pipeline}
\end{figure*}

\subsection{Problem Formulation}

We consider a source domain $\mathcal{D}_S$ of annotated
human-operated gripper demonstrations and a target domain
$\mathcal{D}_T$ of robot teleoperation demonstrations. Both
domains share an action vocabulary
$\mathcal{Y}=\{1,\ldots,C\}$, with $C=22$, but differ in
visual appearance, manipulator embodiment, and execution timing.
Given an untrimmed target-domain video
$\mathbf{V}=\{f_1,\ldots,f_T\}$, our goal is to predict a
frame-level action sequence
$\hat{\mathbf{y}}=\{\hat{y}_1,\ldots,\hat{y}_T\}$.
We first transfer a temporal segmentation model from human
to robot demonstrations, then selectively revise its predicted
segments using sequence-level context. A separate conformal
module represents frame-level uncertainty through prediction
sets $\mathcal{C}(f_t)\subseteq\mathcal{Y}$.
Figure~\ref{fig:pipeline} summarizes the framework.

\subsection{Feature Extraction}

We extract visual features using
X3D-M~\cite{feichtenhofer2020x3d}, pretrained on
Kinetics-400~\cite{carreira2017quo}. For each frame $f_t$,
a 16-frame temporal clip centered at $f_t$ is processed by
the frozen feature extractor to obtain
$\mathbf{x}_t\in\mathbb{R}^{192}$, giving a video
representation $\mathbf{X}\in\mathbb{R}^{192\times T}$.
The feature extractor remains fixed throughout source-domain
pretraining and robot-domain fine-tuning, isolating transfer
to the temporal model. Because centered clips are used, the
evaluated setting is offline rather than strictly causal.

\subsection{Cross-Domain Temporal Action Segmentation}
\label{sec:transfer}

We use MS-TCN++~\cite{li2020mstcnpp} to map visual feature
sequences to frame-level action probabilities. Its multi-stage
temporal model produces $\mathbf{p}^{(s)}_t\in[0,1]^C$ at
stage $s$, with final prediction
\begin{equation}
    \hat{y}_t
    =
    \arg\max_{c\in\mathcal{Y}}p^{(S)}_{t,c}.
\end{equation}

Training uses intermediate supervision at each stage:
\begin{equation}
    \mathcal{L}
    =
    \sum_{s=1}^{S}
    \left(
        \mathcal{L}^{(s)}_{\mathrm{cls}}
        +
        \lambda_{\mathrm{sm}}\mathcal{L}^{(s)}_{\mathrm{sm}}
    \right),
    \label{eq:loss}
\end{equation}
where $\mathcal{L}^{(s)}_{\mathrm{cls}}$ is mean-reduced
cross-entropy with inverse-frequency class weights and label
smoothing of 0.1; frames with annotation sentinel $-1$ are
excluded using \texttt{ignore\_index}. The temporal smoothing
term is
\begin{equation}
    \mathcal{L}^{(s)}_{\mathrm{sm}}
    =
    \frac{1}{|\mathcal{P}|C}
    \sum_{(t,c)\in\mathcal{P}\times\{1,\ldots,C\}}
    \min\left(\delta^{(s)}_{t,c},\tau_{\mathrm{sm}}\right)^2,
    \label{eq:smoothing}
\end{equation}
where
\begin{equation}
    \delta^{(s)}_{t,c}
    =
    \left|
        \log p^{(s)}_{t,c}
        -
        \log p^{(s)}_{t-1,c}
    \right|.
\end{equation}
Here, $\mathcal{P}$ contains adjacent frame pairs for which
both frames have usable labels, preventing smoothing across
unlabeled gaps. We use $\lambda_{\mathrm{sm}}=0.15$ and
$\tau_{\mathrm{sm}}=4.0$.

Let $\mathcal{D}^{\mathrm{tr}}_S$ and
$\mathcal{D}^{\mathrm{tr}}_T$ denote the source and target
training splits. Source pretraining learns
\begin{equation}
    \boldsymbol{\theta}_S^*
    =
    \arg\min_{\boldsymbol{\theta}}
    \sum_{(\mathbf{X},\mathbf{y})
    \in\mathcal{D}^{\mathrm{tr}}_S}
    \mathcal{L}(\mathbf{X},\mathbf{y};\boldsymbol{\theta}),
    \label{eq:source}
\end{equation}
and initializes robot-domain fine-tuning:
\begin{equation}
    \begin{aligned}
        \boldsymbol{\theta}_T^*
        &=
        \arg\min_{\boldsymbol{\theta}}
        \sum_{(\mathbf{X},\mathbf{y})
        \in\mathcal{D}^{\mathrm{tr}}_T}
        \mathcal{L}(\mathbf{X},\mathbf{y};\boldsymbol{\theta}),\\
        \boldsymbol{\theta}^{(0)}
        &=\boldsymbol{\theta}_S^*.
    \end{aligned}
    \label{eq:finetune}
\end{equation}

Human-to-robot and robot-only models use identical target
training data, architecture, and optimization; only
initialization differs, isolating the contribution of
source-domain temporal pretraining. At evaluation, the
temporal model uses an all-ones inference mask, so
ground-truth annotation boundaries do not determine its
activations, and videos are processed and scored separately.

\subsection{Reliability-Aware Selective Sequence Correction}
\label{sec:corrector}

The correction module follows a detect--propose--select design
on frozen backbone predictions: a reliability model flags
potentially incorrect segments, a sequence-aware Transformer
generates replacement proposals, and a frozen gate selectively
accepts eligible revisions.

\noindent\textbf{Predicted segments.}
We partition $\hat{\mathbf{y}}$ into maximal contiguous segments
\begin{equation}
    \mathcal{S}=\{(a_j,b_j,y_j)\}_{j=1}^{J},
\end{equation}
where $a_j$ and $b_j$ are inclusive frame indices and $y_j$
is the backbone label.

\noindent\textbf{Reliability-based flagging.}
A logistic-regression reliability model estimates whether a
predicted segment is incorrect using three prediction-derived
features: segment-mean top-1 confidence, empirical per-class
error rate estimated from validation predictions, and segment
duration normalized by a per-class expected duration. Let
$\mathbf{z}_j\in\mathbb{R}^{3}$ denote this feature vector.
After feature scaling,
\begin{equation}
    F_j
    =
    \mathbf{1}\left[
        \sigma\left(
            \mathbf{w}^{\top}
            \operatorname{Scale}(\mathbf{z}_j)+b
        \right)
        \geq\theta_{\mathrm{flag}}
    \right],
    \label{eq:flag}
\end{equation}
with $\theta_{\mathrm{flag}}=0.39$. The training target is
$\mathbf{1}[y_j\neq\bar y_j]$, where $\bar y_j$ is the
majority ground-truth label over $[a_j,b_j]$; segments with no
labeled frames are excluded. The model, scaler, and threshold
are fit on validation data using demonstration-grouped
cross-validation and frozen before test evaluation.

\noindent\textbf{Sequence-aware proposals.}
Each predicted segment is represented by
\begin{equation}
    \mathbf{h}_j
    =
    [\bar{\mathbf{p}}_j,c_j,m_j,e_j,d_j]
    \in\mathbb{R}^{26},
\end{equation}
where $\bar{\mathbf{p}}_j\in\mathbb{R}^{22}$ is the
segment-mean backbone softmax vector, $c_j$ is top-1
confidence, $m_j$ is the top-two-class margin, $e_j$ is
softmax entropy, and $d_j$ is segment duration normalized by
video length. No predicted segment is dropped.

A Transformer~\cite{vaswani2017attention} processes the
\emph{complete} predicted segment sequence using full,
non-causal self-attention:
\begin{equation}
    (\mathbf{q}_1,\ldots,\mathbf{q}_J)
    =
    g_{\boldsymbol{\phi}}
    (\mathbf{h}_1,\ldots,\mathbf{h}_J).
\end{equation}
The encoder has two pre-norm layers with
$d_{\mathrm{model}}=64$, four attention heads, feedforward
dimension 128, GELU activation, dropout 0.3, fixed sinusoidal
positional encoding, a linear input projection, and a linear
classification head. It is trained with per-segment
cross-entropy using AdamW (learning rate $10^{-3}$, weight
decay $10^{-4}$) for up to 100 epochs with early-stopping
patience 10. Segments without labeled frames remain as context
but are excluded from the loss using
\texttt{ignore\_index}$=-1$.

Training mirrors the backbone: the Transformer is first trained
on human-domain predicted segment sequences with a held-out
human validation split, then fine-tuned on robot-domain
training sequences using an internal demonstration-grouped
80/20 split for early stopping. Robot validation and test
splits are never used for Transformer training.

For every segment,
\begin{equation}
    y_j^{\mathrm{prop}}
    =
    \arg\max_c q_{j,c},
    \qquad
    \pi_j=q_{j,y_j^{\mathrm{prop}}}.
\end{equation}
Proposals are computed for all segments because the Transformer
reasons over the complete sequence; the reliability flag
determines which proposals are eligible to replace a backbone
label.

\noindent\textbf{Selective acceptance.}
Proposals are evaluated from left to right. Previously
processed segments use their resolved labels, while unprocessed
segments retain their original backbone labels. Let $\ell_j$
and $r_j$ denote the neighboring labels available when
processing segment $j$. Define local transition compatibility
as
\begin{equation}
    B_j(c)
    =
    \log P_{\mathrm{tr}}(c\mid\ell_j)
    +
    \log P_{\mathrm{tr}}(r_j\mid c),
\end{equation}
omitting the corresponding term when a neighbor is absent.
The bigram prior $P_{\mathrm{tr}}$ is estimated from
consecutive ground-truth segment labels in robot training
videos using add-$\alpha$ Laplace smoothing ($\alpha=0.5$)
and frozen before evaluation. A class never observed as a
transition's first element defaults to a uniform outgoing
distribution.

The transition-score change is
\begin{equation}
    \Delta_j
    =
    B_j(y_j^{\mathrm{prop}})
    -
    B_j(y_j),
    \label{eq:transition_delta}
\end{equation}
and a changed proposal is accepted when
\begin{equation}
    \begin{aligned}
        A_j
        ={}&
        F_j\,
        \mathbf{1}[y_j^{\mathrm{prop}}\neq y_j]\\
        &{}\cdot
        \mathbf{1}[\pi_j\geq\tau_{\mathrm{gate}}]\,
        \mathbf{1}[\Delta_j\geq m_{\mathrm{gate}}],
    \end{aligned}
    \label{eq:gate}
\end{equation}
giving
\begin{equation}
    \tilde{y}_t
    =
    \begin{cases}
        y_j^{\mathrm{prop}}, & A_j=1,\\
        y_j, & A_j=0,
    \end{cases}
    \quad a_j\leq t\leq b_j.
    \label{eq:reconstruction}
\end{equation}
Relabeling may merge adjacent equal-label segments but cannot
introduce a new boundary inside an original predicted segment.

We use $m_{\mathrm{gate}}=0$ for both backbones, with
$\tau_{\mathrm{gate}}=0.7$ for human-to-robot and
$\tau_{\mathrm{gate}}=0.5$ for robot-only. These are selected
on robot validation sequences only: candidate configurations
must preserve or improve validation Edit relative to the
corresponding MS-TCN++ baseline for all three seeds, after
which the configuration maximizing mean validation frame
accuracy is frozen and shared across seeds.

For the raw Transformer ablation, all changed proposals are
applied directly without reliability-based eligibility,
confidence thresholding, or transition-compatibility
filtering, isolating the effect of selective acceptance.

\noindent\textbf{Deployment validity.}
All corrector inputs derive from backbone outputs and predicted
segment geometry; no ground-truth boundaries or
ground-truth-derived features are used at inference. Ground
truth enters only during fitting of the reliability model,
Transformer, and transition prior. The reliability model,
scaler, flag threshold, bigram prior, and gate configuration
are all frozen before test evaluation; no test label or test
metric influences them.

\begin{table*}[t]
\centering
\caption{
Temporal backbone selection on source-domain human validation data
(upper) and sequence-aware corrector selection on robot validation data
(lower). Correctors are pretrained on source-domain predictions and
fine-tuned on target-domain robot demonstrations, and apply every
proposed relabeling (no selective gate).
}
\label{tab:model_selection}
\footnotesize
\setlength{\tabcolsep}{3pt}
\renewcommand{\arraystretch}{1.09}

\begin{tabular*}{\textwidth}{@{\extracolsep{\fill}}lccccc@{}}
\toprule
Method & Edit & F1@10 & F1@25 & F1@50 & Acc \\
\midrule

\multicolumn{6}{@{}l}{\textit{Source-domain model}} \\
\addlinespace[1pt]

TCN
& 3.10 & 3.60 & 3.10 & 2.20 & 41.96 \\

ASFormer
& 21.90 & 27.20 & 24.10 & 13.90 & 34.48 \\

BiLSTM
& 24.24 & 19.80 & 14.50 & 6.20 & 20.70 \\

MS-TCN++ 10L
& $29.1 \pm 1.7$ & $35.9 \pm 1.4$ & $31.7 \pm 1.2$
& $25.2 \pm 0.5$ & $57.1 \pm 1.0$ \\

MS-TCN++ 12L
& $56.7 \pm 1.4$ & $57.2 \pm 2.6$ & $50.2 \pm 3.6$
& $35.3 \pm 1.5$ & $60.8 \pm 2.4$ \\

\textbf{MS-TCN++ 14L} (backbone)
& $\mathbf{70.3} \pm 3.3$
& $\mathbf{65.4} \pm 2.1$
& $\mathbf{58.7} \pm 2.6$
& $\mathbf{43.1} \pm 2.0$
& $\mathbf{68.8} \pm 0.2$ \\

\midrule

\multicolumn{6}{@{}l}{
\textit{Sequence-aware transformer}
} \\
\addlinespace[1pt]

MS-TCN baseline
& $69.59 \pm 0.96$ & $34.88 \pm 1.69$ & $27.36 \pm 0.84$
& $16.89 \pm 0.80$ & $31.81 \pm 0.18$ \\

BiLSTM-CRF
& $66.45 \pm 3.94$ & $34.84 \pm 1.82$ & $26.94 \pm 2.08$
& $16.80 \pm 0.78$ & $31.19 \pm 1.68$ \\

ASFormer
& $69.91 \pm 4.85$ & $37.77 \pm 1.85$ & $28.76 \pm 0.82$
& $17.97 \pm 1.05$ & $32.91 \pm 0.34$ \\

Transformer encoder (Main)
& $\mathbf{68.79} \pm 3.22$ & $\mathbf{37.01} \pm 2.03$
& $\mathbf{28.31 }\pm 1.39$ & $\mathbf{17.91} \pm 1.04$
& $\mathbf{32.33} \pm 1.16$ \\

Simple Corrector$^\ddagger$
& 61.73 & 34.71 & 27.17 & 16.37 & 30.66 \\

\bottomrule
\end{tabular*}

\vspace{1pt}

{\scriptsize
MS-TCN++ 14L is selected as the temporal backbone, and the Transformer encoder
is selected as the sequence-aware corrector. Correctors operate on
predictions from the Human$\rightarrow$Robot MS-TCN++ 14L backbone.
Mean~$\pm$~std over seeds 42/123/456; $^\ddagger$Simple Corrector is
reported for a single seed.
\par}

\end{table*}
\subsection{Uncertainty Quantification with Conformal Prediction}
\label{sec:cp}

We use CP~\cite{DBLP:journals/corr/abs-2107-07511,
stutz2021learning} to construct frame-level sets of plausible
action labels. Coverage and prediction-set size are reported
for the baseline and corrected pipelines.

For the corrected pipeline, CP is calibrated against the
probability stream produced by the deployed system. Frames
inside an accepted segment ($A_j=1$) use the Transformer's
segment-level softmax vector, constant across that segment;
all other frames retain the backbone's per-frame softmax
vector. Temperature scaling and conformal calibration are fit
independently for the baseline and corrected probability
streams.

\noindent\textbf{Temperature scaling.}
For logits $\mathbf{z}_t$ from a given probability-producing
pipeline, temperature-scaled probabilities are
\begin{equation}
    \bar{p}_t(c)
    =
    \frac{\exp(z_{t,c}/T_{\mathrm{temp}})}
    {\sum_{c'=1}^{C}\exp(z_{t,c'}/T_{\mathrm{temp}})},
\end{equation}
where $T_{\mathrm{temp}}>0$ is fitted by minimizing validation
negative log-likelihood~\cite{guo2017calibration} and fixed
before conformal calibration.

\noindent\textbf{Global split-conformal prediction.}
For calibration frames,
\begin{equation}
    s_i=1-\bar{p}_i(y_i),
    \qquad i\in\mathcal{I}_{\mathrm{cal}},
\end{equation}
where $n=|\mathcal{I}_{\mathrm{cal}}|$. For target
miscoverage $\alpha$,
\begin{equation}
    k
    =
    \min\left(
        n,\left\lceil(n+1)(1-\alpha)\right\rceil
    \right),
    \qquad
    \hat q_{\mathrm{global}}=s_{(k)},
    \label{eq:global_quantile}
\end{equation}
where $s_{(k)}$ is the $k$th smallest calibration score.
The global prediction set is
\begin{equation}
    \mathcal{C}_{\mathrm{global}}(f_t)
    =
    \left\{
        c\in\mathcal{Y}:
        \bar p_t(c)\geq1-\hat q_{\mathrm{global}}
    \right\}.
\end{equation}

When $\lceil(n+1)(1-\alpha)\rceil>n$, the implementation
clips the rank to $n$ and uses the largest observed
calibration score. The standard finite-sample coverage
guarantee does not generally apply at such operating points.

\noindent\textbf{Class-conditional prediction.}
For class $c$, define
\begin{equation}
    \mathcal{S}_c
    =
    \{s_i:i\in\mathcal{I}_{\mathrm{cal}},\,y_i=c\},
    \qquad
    n_c=|\mathcal{S}_c|.
\end{equation}
For $n_c\geq n_{\min}$, with $n_{\min}=30$, we compute
$\hat q_c$ using the same order-statistic rule as
Eq.~\ref{eq:global_quantile}. For classes with fewer than
$n_{\min}$ calibration examples, including classes absent
from calibration, we use
$\hat q_c=\hat q_{\mathrm{global}}$. The resulting set is
\begin{equation}
    \mathcal{C}_{\mathrm{class}}(f_t)
    =
    \left\{
        c\in\mathcal{Y}:
        \bar p_t(c)\geq1-\hat q_c
    \right\}.
\end{equation}
Low-support or absent classes therefore receive no
class-specific calibration guarantee under this rule.

\noindent\textbf{Interpretation.}
Formal conformal coverage requires exchangeability, which
physical-demonstration separation does not ensure for temporally
correlated frames. We therefore report empirical coverage and mean
set size on held-out robot videos. Prediction sets characterize
label ambiguity, while the correction gate decides whether a
relabeling is applied.

\begin{table*}[t]
\centering
\caption{
Target-domain test performance over three seeds. Raw Transformer applies all
changed proposals; Selective Gate uses frozen reliability, confidence,
and transition-compatibility criteria.
}
\label{tab:robot_transfer}
\footnotesize
\setlength{\tabcolsep}{3pt}
\renewcommand{\arraystretch}{1.1}

\begin{tabular*}{\textwidth}{@{\extracolsep{\fill}}lccccc@{}}
\toprule
Method & Edit & F1@10 & F1@25 & F1@50 & Acc \\
\midrule

Human-Only (No Robot Adaptation)
& $37.71 \pm 1.97$ & $18.70 \pm 1.01$ & $11.68 \pm 0.21$
& $4.33 \pm 0.62$ & $13.40 \pm 0.20$ \\

\addlinespace[1pt]

Robot-Only MS-TCN++
& $55.23 \pm 5.96$ & $32.23 \pm 2.28$ & $24.01 \pm 1.77$
& $10.97 \pm 0.28$ & $28.48 \pm 1.26$ \\

\quad + Raw Transformer
& $40.53 \pm 5.90$ & $32.40 \pm 3.04$ & $24.56 \pm 2.44$
& $9.75 \pm 1.22$ & $29.87 \pm 0.83$ \\

\quad + Selective Gate
& $55.99 \pm 4.25$
& $32.99 \pm 1.59$ & $24.69 \pm 1.48$
& $11.25 \pm 0.66$ & $28.82 \pm 0.94$ \\

\addlinespace[1pt]

Human$\rightarrow$Robot MS-TCN++
& $72.51 \pm 2.19$ & $38.52 \pm 1.72$ & $30.81 \pm 1.33$
& $16.86 \pm 0.46$ & $34.11 \pm 0.55$ \\

\quad + Raw Transformer
& $67.86 \pm 4.11$ & $39.59 \pm 4.28$ & $31.39 \pm 2.47$
& $17.19 \pm 0.99$ & $34.11 \pm 1.84$ \\

\quad + Selective Gate
& $\mathbf{73.45} \pm 2.81$
& $\mathbf{38.58} \pm 1.83$
& $\mathbf{30.83} \pm 1.31$
& $\mathbf{16.89} \pm 0.58$
& $\mathbf{34.19} \pm 0.60$ \\

\bottomrule
\end{tabular*}
\vspace{1pt}
{\scriptsize
Human-Only: source-domain training without robot adaptation.
Robot-Only: target-domain training only.
Human$\rightarrow$Robot: source-domain pretraining followed by target-domain
fine-tuning. All results are mean$\pm$std over seeds 42/123/456.
\par}
\end{table*}

\section{Experiments}
\label{sec:experiments}
\vspace{-.4em}

\begin{figure}[t]
\centering
\includegraphics[width=\linewidth]{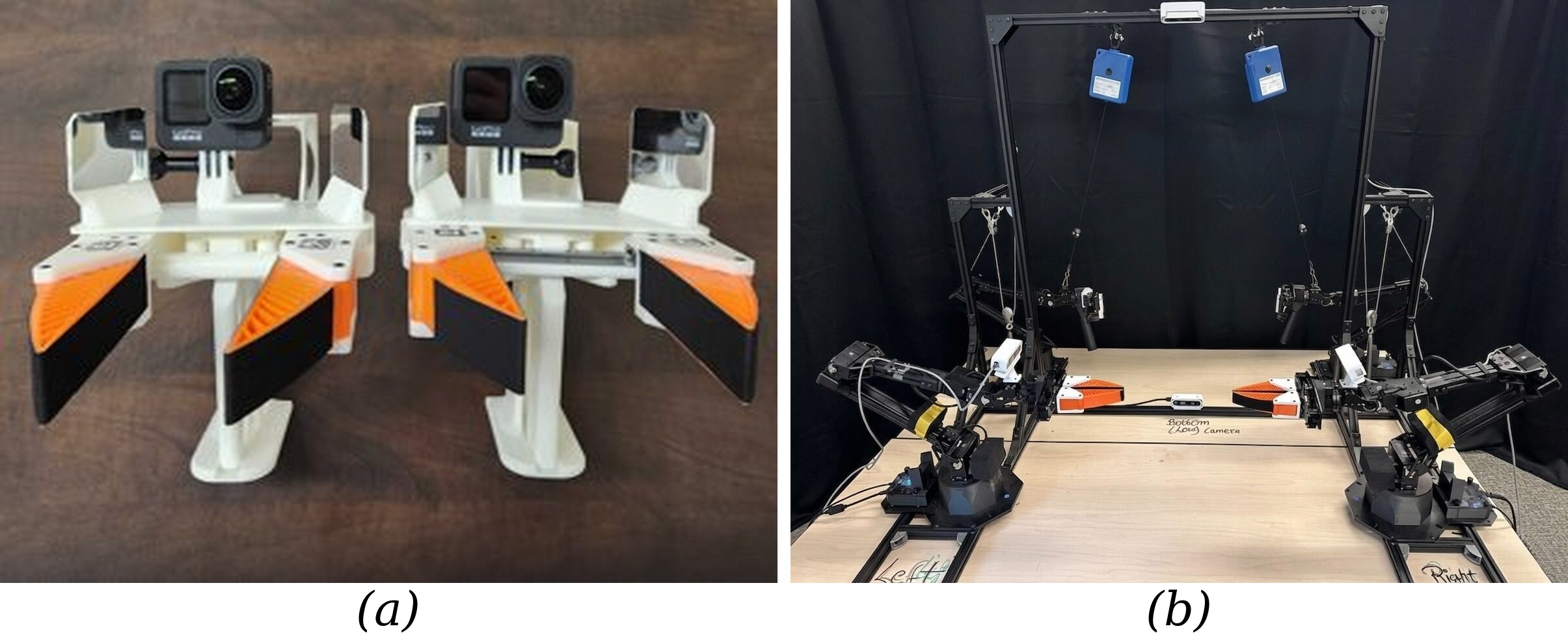}
\caption{
\textbf{Hardware platforms.}
(a) UMI human-operated gripper platform used for source-domain
demonstrations.
(b) ALOHA bimanual robot platform used for target-domain
teleoperation demonstrations.
}
\label{fig:platforms}
\vspace{-1em}
\end{figure}

\noindent\textbf{Data Collection and Annotation.}
We recorded toy-car assembly demonstrations across two platforms as shown in
Fig.~\ref{fig:platforms}, covering picking, placing, inserting,
screw-driving, fastening, and inspection.

UMI~\cite{chi2024umi} is used to collect human-operated gripper
demonstrations in the source domain, while ALOHA~\cite{zhao2023aloha}
is used to collect robot teleoperation demonstrations in the target
domain. Although both domains perform the same overall toy-car
assembly task, executions vary in action ordering, segment duration,
manipulation strategy, and task progression. After collapsing each
physical demonstration to its ordered segment-label sequence, the
human training split contains 36 distinct sequences across 42
demonstrations, while the ALOHA training, validation, and test splits
contain 5/9, 9/13, and 12/17 distinct sequences/demonstrations,
respectively.

Demonstrations were collected across different environmental
configurations and performed by five operators, with robot-domain
recordings captured using both GoPro and RealSense cameras. This
introduces variability in execution speed, motion style, object
handling, and action timing. Each demonstration is annotated at the
segment level using a shared vocabulary of 22 fine-grained action
classes (Fig.~\ref{fig:dataset_segmentation}), enabling direct
human-to-robot transfer without label remapping.

The human training set comprises 42 physical demonstrations
(84 videos). Robot splits comprise 9/13/15/8/17 physical demonstrations
(18/26/30/16/34 videos) for training, validation, conformal
calibration, conformal development, and testing, respectively.
Each physical demonstration is recorded from left- and right-camera
views, which are treated as separate videos and processed independently
without multi-view fusion. The nine-demonstration budget refers
specifically to backbone training. Splits are disjoint at the
physical-demonstration level, with both camera views from a
demonstration assigned to the same split.

\noindent\textbf{Experimental Setup.}
All experiments use precomputed features from a frozen
X3D-M~\cite{feichtenhofer2020x3d} backbone pretrained on
Kinetics-400, producing a 192-dimensional feature vector per frame
from a centered 16-frame clip. We evaluate MS-TCN++ variants with
10, 12, and 14 layers per stage, each using 4 stages, 64 feature maps,
and 22 output classes. The selected 14-layer model is used for all
robot-domain experiments and trained using the protocol in
Section~\ref{sec:transfer}. Experiments are performed on a single
NVIDIA A100 GPU.

We report frame-level accuracy (Acc), normalized edit score (Edit),
and segmental F1 at temporal IoU thresholds of $10\%$, $25\%$, and
$50\%$. Edit compares collapsed action sequences using normalized
Levenshtein distance, whereas segmental F1 evaluates overlap between
predicted and ground-truth action segments. Results are reported as
mean~$\pm$~std over three random seeds (42, 123, 456).

\noindent\textbf{Baselines.}
Robot-Only trains MS-TCN++ from random initialization on the robot
training split. Human$\rightarrow$Robot initializes from the
human-pretrained checkpoint and fine-tunes on the same robot split
using an identical optimization protocol. Thus, the two conditions
differ only in temporal-backbone initialization, isolating the
effect of human-domain temporal pretraining.

\section{Results}
\vspace{-.4em}

\noindent\textbf{Architecture Selection.}
Table~\ref{tab:model_selection} compares candidate temporal backbones
on human validation data. TCN attains $41.96\%$ frame accuracy but
only $3.10$ Edit, indicating poor sequence-level segmentation.
ASFormer and BiLSTM reach $21.90$ and $24.24$ Edit, respectively.
Increasing MS-TCN++ depth from 12 to 14 layers improves accuracy from
$60.8{\pm}2.4\%$ to $68.8{\pm}0.2\%$ and Edit from
$56.7{\pm}1.4$ to $70.3{\pm}3.3$. We therefore use
\textbf{MS-TCN++ 14L} for all robot-domain experiments.
 
The lower block compares ungated 26-D sequence correctors on robot
validation data. ASFormer and the Transformer perform comparably:
per-seed outcomes are split, and their $1.1$-point mean Edit
difference is smaller than either model's seed-to-seed standard
deviation. We use the Transformer, which has approximately $70$K
parameters compared with ASFormer's $172$K. The Transformer,
BiLSTM-CRF, and the Simple Corrector all reduce mean Edit relative
to the uncorrected backbone, and ASFormer improves it by only
$0.32$. Because these rows apply every proposed relabeling, they
characterize proposal models without selective acceptance rather
than the deployed correction pipeline.

\noindent\textbf{Human-to-Robot Transfer.}
Table~\ref{tab:robot_transfer} reports robot test performance over
three seeds. Without robot adaptation, the human-only model achieves
$13.40{\pm}0.20\%$ frame accuracy, indicating a substantial
source--target domain gap. Robot-Only MS-TCN++ reaches
$28.48{\pm}1.26\%$ accuracy and $55.23{\pm}5.96$ Edit, whereas
human-domain pretraining followed by robot fine-tuning reaches
$34.11{\pm}0.55\%$ accuracy and $72.51{\pm}2.19$ Edit, improvements
of $+5.63$ and $+17.28$ points under identical robot training data.
F1 also improves at all three overlap thresholds. These results show
that human-domain pretraining provides a useful initialization for
robot-domain temporal segmentation despite differences in embodiment
and visual appearance.
\begin{table}[t]
\centering
\caption{Reliability analyzer performance for erroneous-segment
detection (mean$\pm$std over three seeds).}
\label{tab:reliability_analyzer}
 
\footnotesize
\setlength{\tabcolsep}{4pt}
\renewcommand{\arraystretch}{1.05}
 
\begin{tabular*}{\columnwidth}
{@{\extracolsep{\fill}}lccc@{}}
\toprule
Metric & HAND & H$\rightarrow$R & Robot-Only \\
\midrule
Error prevalence
& $.568{\pm}.016$
& $.680{\pm}.015$
& $.724{\pm}.026$ \\
 
\midrule
AUROC
& $.674{\pm}.039$
& $.755{\pm}.018$
& $.765{\pm}.046$ \\
 
AUPRC
& $.694{\pm}.038$
& $.859{\pm}.025$
& $.880{\pm}.038$ \\
 
\midrule
 
Precision
& $.635{\pm}.034$
& $.777{\pm}.020$
& $.820{\pm}.032$ \\
 
Recall
& $.834{\pm}.012$
& $.817{\pm}.013$
& $.793{\pm}.081$ \\
 
F1
& $.721{\pm}.024$
& $.796{\pm}.017$
& $.806{\pm}.055$ \\
 
\bottomrule
\end{tabular*}
 
\vspace{1.2pt}
 
{\scriptsize
HAND: validation (out-of-fold); H$\rightarrow$R, Robot-Only: test.
Error prevalence (fraction of mislabeled segments) is the chance
level for precision and AUPRC.
\par}
\vspace{-1em}
\end{table}
The larger improvement in Edit than in frame accuracy suggests that
source pretraining particularly benefits sequence-level temporal
structure, such as action ordering and duration patterns, while
frame-level recognition remains challenging after transfer. Because
the domains share an action vocabulary, however, these results do
not establish embodiment-invariant visual representations.


\begin{figure*}[t]
    \centering
    \begin{subfigure}[t]{0.49\textwidth}
        \centering
        \includegraphics[width=\linewidth,height=0.20\textheight]{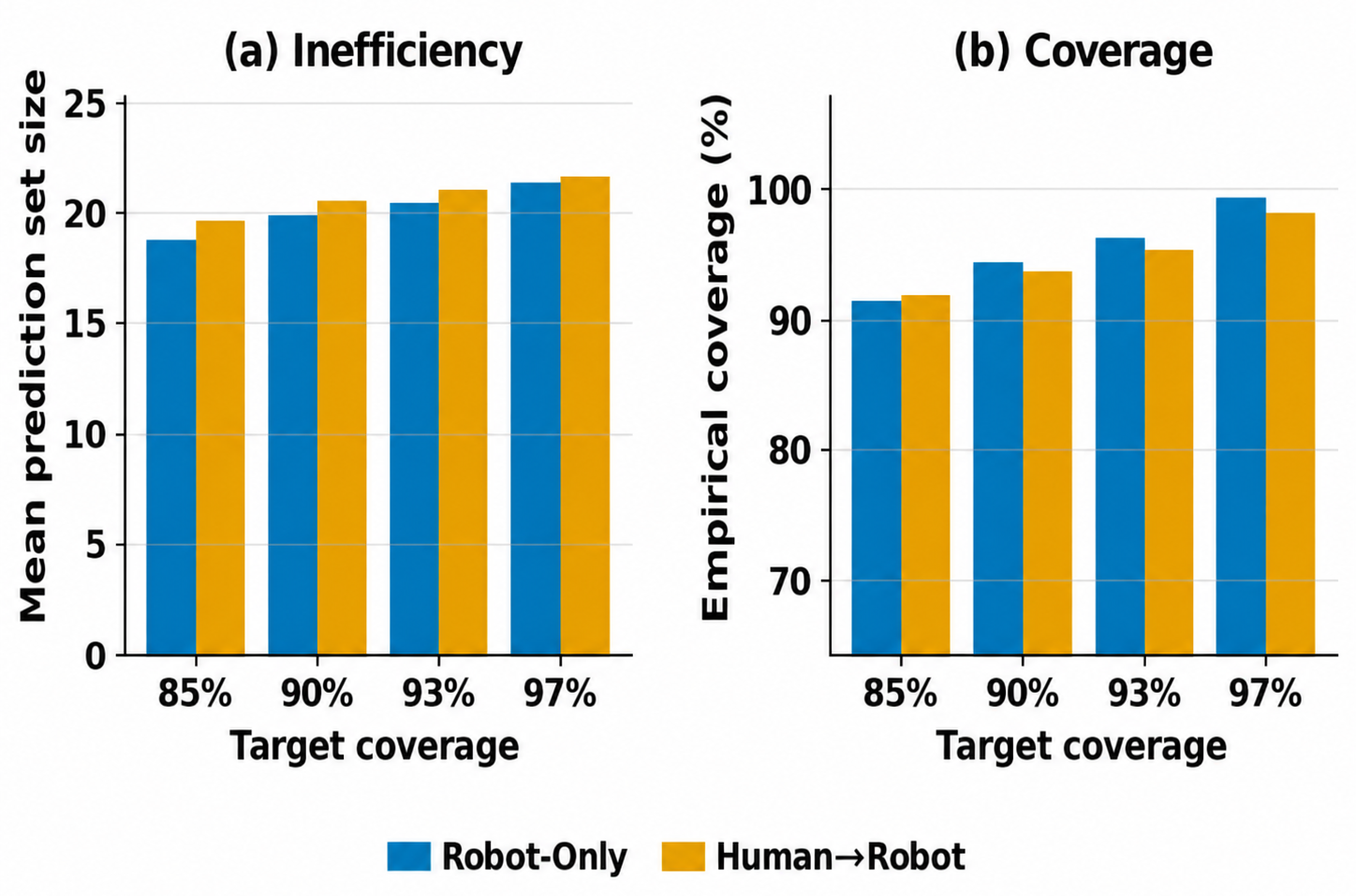}
        \caption{Global split-conformal prediction.}
        \label{fig:cp_standard}
    \end{subfigure}
    \hfill
    \begin{subfigure}[t]{0.49\textwidth}
        \centering
        \includegraphics[width=\linewidth,height=0.20\textheight]{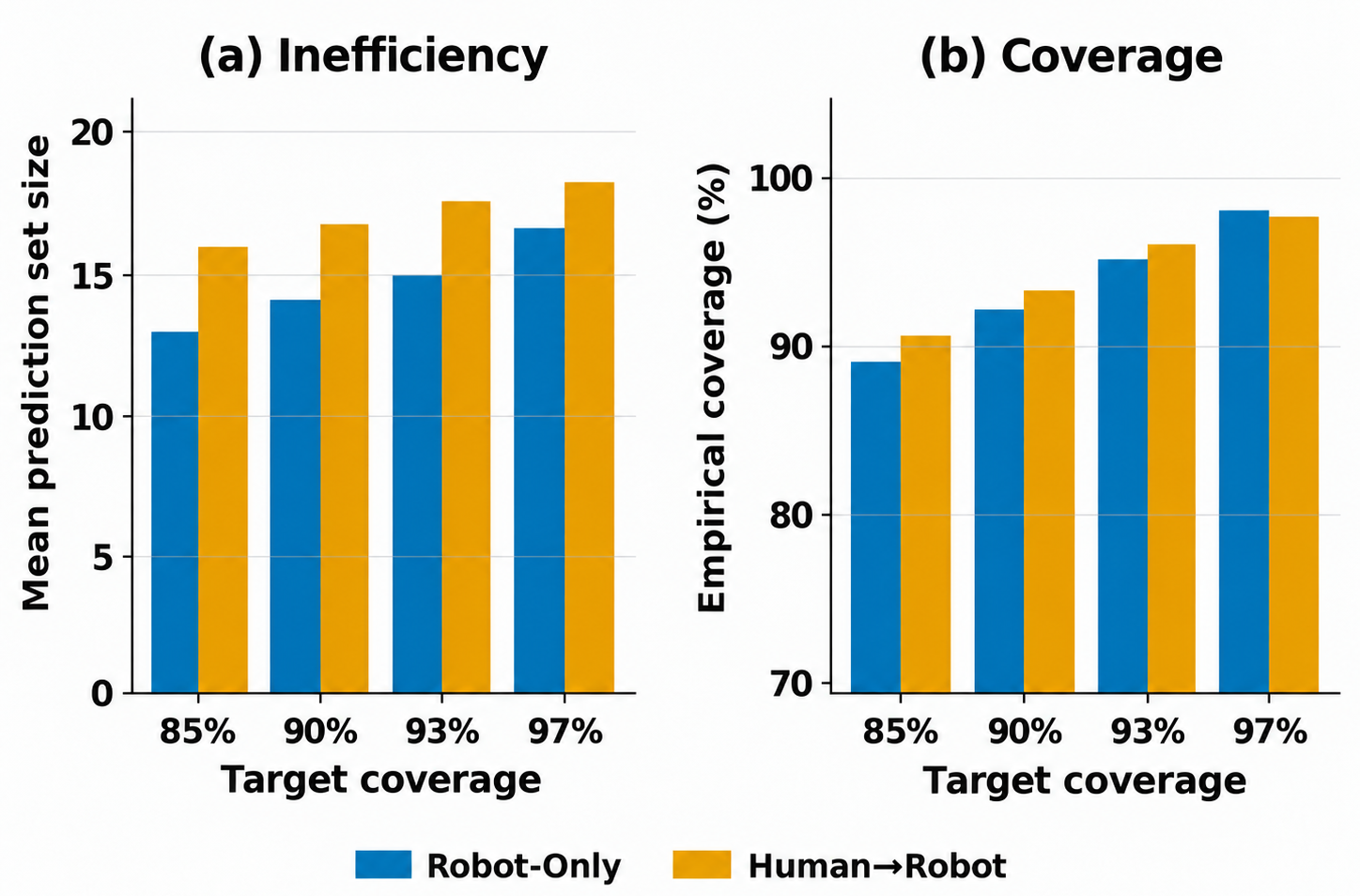}
        \caption{Class-conditional prediction.}
        \label{fig:cp_class}
    \end{subfigure}
    \caption{
    CP results on the robot test set for Robot-Only and
    Human$\rightarrow$Robot using (a) global split-conformal prediction
    and (b) class-conditional thresholds with minimum-support fallback.
    Each panel reports empirical coverage and mean prediction-set size
    at target levels of 85, 90, 93, and 97\%, averaged over three seeds.
    }
    \label{fig:cp_bars}
\end{figure*}

\noindent\textbf{Reliability Analyzer.}
Table~\ref{tab:reliability_analyzer} evaluates the analyzer as a
binary classifier for incorrect backbone segments using the frozen
threshold $\theta_{\mathrm{flag}}=0.39$. On the
Human$\rightarrow$Robot test split, it achieves
$0.755{\pm}0.018$ AUROC, $0.777{\pm}0.020$ precision,
$0.817{\pm}0.013$ recall, and $0.796{\pm}0.017$ F1.
Robot-Only exhibits similar discrimination, with
$0.765{\pm}0.046$ AUROC and $0.806{\pm}0.055$ F1. On HAND
validation, AUROC and F1 are $0.674{\pm}0.039$ and
$0.721{\pm}0.024$, respectively. Erroneous segments are common: error prevalence is $0.680$ for
Human$\rightarrow$Robot and $0.724$ for Robot-Only on test, and
$0.568$ on HAND validation. Precision therefore exceeds its chance
level by only about $0.07$--$0.10$, whereas AUPRC exceeds it by
$0.16$--$0.18$ and AUROC ($0.755$--$0.765$) is well above $0.5$,
indicating that the analyzer ranks erroneous segments meaningfully
above correct ones.
Overall, the analyzer identifies a substantial fraction of incorrect
segments while retaining imperfect precision, motivating selective
treatment of subsequent replacement proposals.

\noindent\textbf{Selective Sequence Correction.}
Unrestricted correction does not reliably improve temporal
segmentation. For Human$\rightarrow$Robot, the Raw Transformer
reduces Edit from $72.51$ to $67.86$ while leaving mean frame
accuracy at $34.11\%$. For Robot-Only, accuracy increases from
$28.48\%$ to $29.87\%$, but Edit decreases substantially from
$55.23$ to $40.53$. Thus, improved frame-level recognition does
not necessarily preserve the collapsed temporal action sequence,
and unrestricted segment relabeling can degrade Edit.

Selective acceptance largely prevents this degradation. It reaches
$73.45{\pm}2.81$ Edit for Human$\rightarrow$Robot and
$55.99{\pm}4.25$ for Robot-Only, compared with backbone scores of
$72.51{\pm}2.19$ and $55.23{\pm}5.96$, respectively. The resulting
changes relative to the backbone are small, indicating that the
current correction pipeline primarily serves as a safeguard against
harmful unrestricted relabeling rather than as a major source of
additional segmentation performance. Together with the analyzer's
relatively high recall, these results suggest that identifying
potentially incorrect segments are more successful than generating
consistently beneficial replacement labels, making proposal quality
a key direction for improvement.


\begin{figure}[t]
  \centering
  \includegraphics[width=\columnwidth]{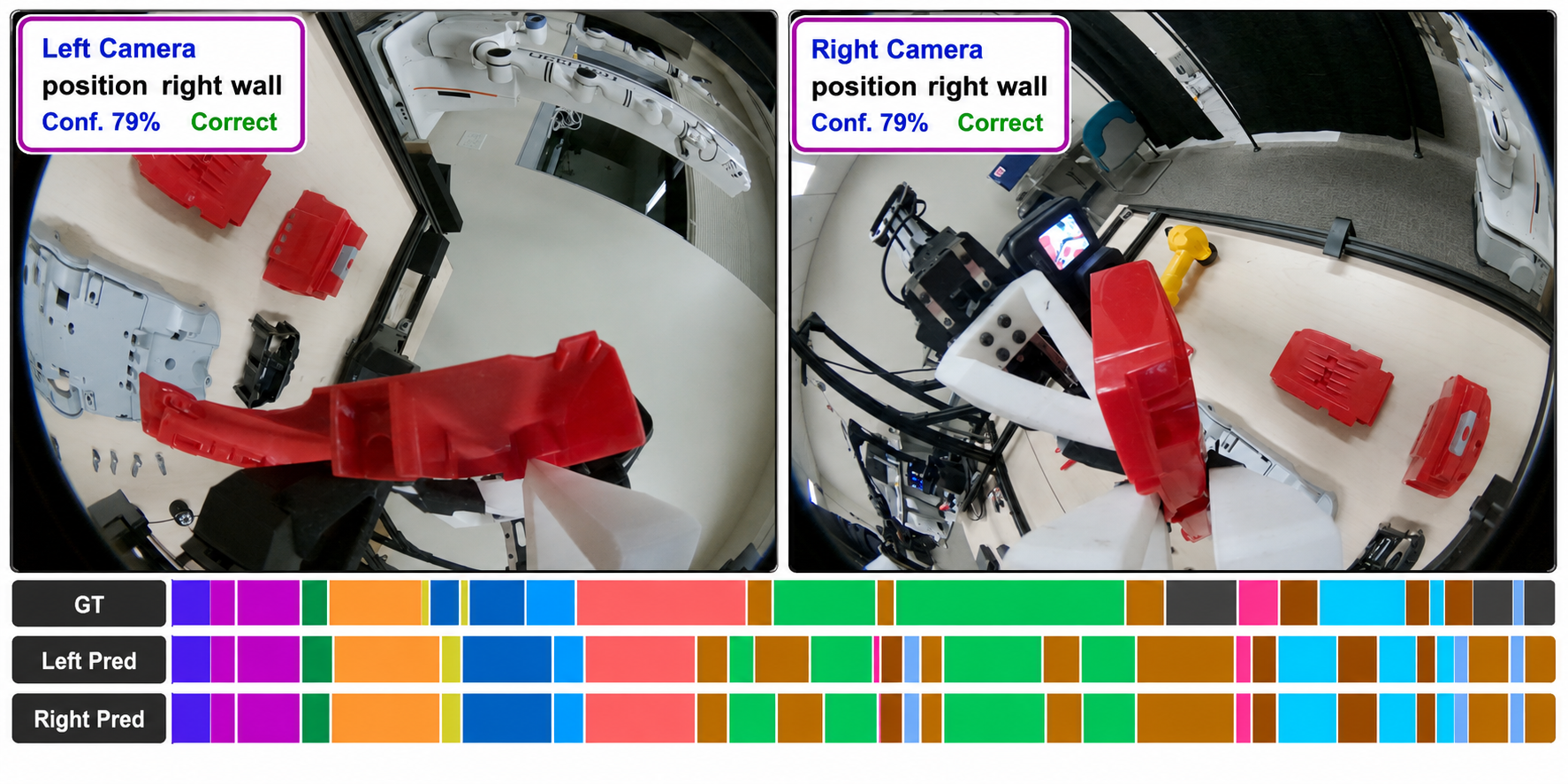}
\caption{
    Robot test sequence from synchronized left and right wrist
    cameras, with ground-truth (GT) and MS-TCN++ predicted segments
    for each view.
  }
  \label{fig:qual_results}
  \vspace{-1.5em}
\end{figure}

\noindent\textbf{Segmentation Visualization.}
Figure~\ref{fig:qual_results} shows representative predictions from
synchronized robot views. The model recovers much of the dominant
action structure, while remaining errors occur near action
transitions and within longer segments containing visually similar
or underrepresented actions.


\noindent\textbf{Uncertainty Quantification.}
We evaluate CP using empirical coverage and mean prediction-set
size on the held-out robot test split, averaged over three seeds
(Fig.~\ref{fig:cp_bars}). For the corrected pipelines at the $93\%$
target, global split CP attains $96.51\%$ empirical coverage with
mean set size $21.05$ for Human$\rightarrow$Robot and $97.39\%$
coverage with set size $20.65$ for Robot-Only. These sets approach
the full 22-class vocabulary and over-cover the target, so they
carry little information about the correct action.
 
Class-conditional thresholds with minimum-support fallback reduce
set size while attaining the target coverage empirically. For
Human$\rightarrow$Robot and Robot-Only, respectively, coverage/set
size is $95.38\%/17.41$ and $95.31\%/14.79$ at the $93\%$ target,
and $97.32\%/18.41$ and $97.80\%/15.99$ at $97\%$. Prediction sets
nevertheless remain large, revealing substantial residual ambiguity.
Thus, high empirical coverage alone should not be interpreted as
high-confidence recognition: the large sets expose substantial
uncertainty among the 22 action classes.
 
At the $93\%$ class-conditional operating point, Robot-Only
correction changes coverage/set size from $94.89\%/14.90$ to
$95.31\%/14.79$. For Human$\rightarrow$Robot, it changes from
$95.33\%/17.19$ to $95.38\%/17.41$. Thus, correction has little
effect on conformal efficiency relative to the backbone probability
stream. Because within-video temporal dependence complicates
exchangeability, these results characterize empirical held-out
coverage rather than establishing formal coverage guarantees.

\section{Conclusion and Limitations}
We presented an uncertainty-aware human-to-robot action segmentation framework that combines temporal transfer, selective correction, and conformal prediction. Pretraining on human demonstrations and fine-tuning on nine robot demonstrations improved test Edit from 55.23 to 72.51 and frame accuracy from 28.48\% to 34.11\% over robot-only training. Selective correction avoided degradation from unrestricted relabeling, while conformal prediction achieved high coverage with large prediction sets. The study is limited to one task and platform, a shared action vocabulary, independently processed camera views, and offline inference; temporal dependence also limits formal coverage claims. Future work will improve recognition with limited robot data, use prediction-set size to guide correction, exploit task structure without training, and support live operator assistance.
\bibliographystyle{IEEEtran}
\bibliography{references}

\end{document}